\documentclass[final]{cvpr}

\usepackage{times}
\usepackage{epsfig}
\usepackage{graphicx}
\usepackage{amsmath}
\usepackage{amssymb}
\usepackage{multirow}
\usepackage{booktabs}

\usepackage{caption}
\usepackage{subfigure}
\usepackage{color}
\usepackage{verbatim}
\usepackage{enumerate}

\usepackage[pagebackref=true,breaklinks=true,colorlinks,bookmarks=false]{hyperref}

\def\cvprPaperID{5174} 
\def\confYear{CVPR 2021}

\begin{document}

\title{Adaptive Class Suppression Loss for Long-Tail Object Detection}

\author{Tong Wang\textsuperscript{1,2}, Yousong Zhu\textsuperscript{1,3}, Chaoyang Zhao\textsuperscript{1}, Wei Zeng\textsuperscript{4,5}, Jinqiao Wang\textsuperscript{1,2,6}, Ming Tang\textsuperscript{1} \\
	\\
\textsuperscript{1} National Laboratory of Pattern Recognition, Institute of Automation, \\ 
Chinese Academy of Sciences, Beijing, China\\
\textsuperscript{2} School of Artificial Intelligence, University of Chinese Academy of Sciences,\\
Beijing, China\\
\textsuperscript{3} ObjectEye Inc., Beijing, China \\
\textsuperscript{4} Peking University, Beijing, China \\
\textsuperscript{5} Peng Cheng Laboratory, Shenzhen, China \\
\textsuperscript{6} NEXWISE Co., Ltd., Guangzhou, China \\
{\tt\small {\{tong.wang,yousong.zhu,chaoyang.zhao,jqwang,tangm\}@nlpr.ia.ac.cn}} \\
{\tt\small {weizeng@pku.edu.cn}}
}

\maketitle
\pagestyle{empty}
\thispagestyle{empty}

\begin{abstract}
	To address the problem of long-tail distribution for the large vocabulary object detection task, existing methods usually divide the whole categories into several groups and treat each group with different strategies. These methods bring the following two problems. One is the training inconsistency between adjacent categories of similar sizes, and the other is that the learned model is lack of discrimination for tail categories which are semantically similar to some of the head categories. In this paper, we devise a novel Adaptive Class Suppression Loss (ACSL) to effectively tackle the above problems and improve the detection performance of tail categories. Specifically, we introduce a statistic-free perspective to analyze the long-tail distribution, breaking the limitation of manual grouping. According to this perspective, our ACSL adjusts the suppression gradients for each sample of each class adaptively, ensuring the training consistency and boosting the discrimination for rare categories. Extensive experiments on long-tail datasets LVIS and Open Images show that the our ACSL achieves 5.18\% and 5.2\% improvements with ResNet50-FPN, and sets a new state of the art. Code and models are available at \href{https://github.com/CASIA-IVA-Lab/ACSL}{https://github.com/CASIA-IVA-Lab/ACSL}. 
	
\end{abstract}

\section{Introduction}
With the advent of deep Convolutional Neural Network, researchers have achieved significant progress on object detection task. Many efforts have been paid to refresh the record on classic benchmarks like PASCAL VOC~\cite{voc} and MS COCO~\cite{coco}. However, these benchmarks usually have limited quantities of classes and exhibit relatively balanced category distribution. Whereas, in real-world scenarios, data usually comes with a long-tail distribution. A few head classes (frequent classes) contribute most of the training samples, while the huge number of tail classes (rare classes) are under-represented in data. Such extremely imbalanced class distribution proposes new challenges for researchers. 
 
\begin{figure}
	\centering
	\includegraphics[width=6.5cm]{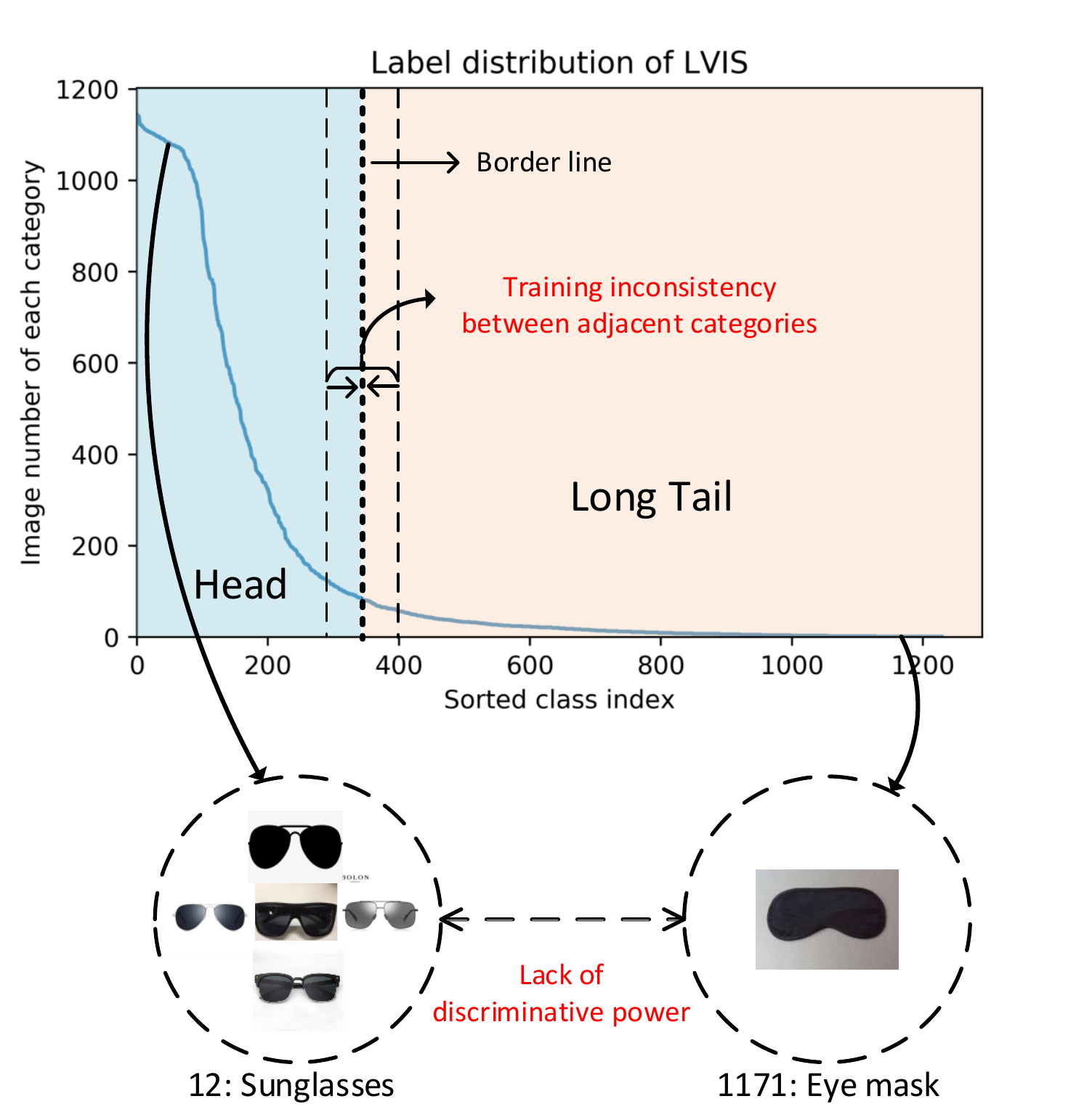}\\
	\caption{The label distribution of LVIS~\cite{lvis} dataset. The x-axis represents the sorted category index of LVIS. The y-axis is the image number of each category.}
	\vspace{-0.5cm}
	\label{fig:distribution}
\end{figure}

An intuitive solution is to re-balance the data distribution by re-sampling technique~\cite{over-sampling, class-imbalance, relay, mahajan2018exploring}. By over-sampling the tail or under-sampling the head classes, a less imbalanced distribution could be artificially generated. Nevertheless, over-sampling usually brings undesirable over-fitting issues on tail classes. And under-sampling may miss valuable information of head classes. The effectiveness of such techniques is limited when the dataset is extremely imbalanced. To improve the performance of tail classes while avoiding the over-fitting issue, Tan \etal ~\cite{eql} devises Equalization Loss, which argues that the poor performance of tail classes originates from the over-suppression of samples from head classes. Since tail classes only contain few samples, they receive much more negative gradients than positive ones during training, thus they are consistently in a state of being suppressed in most of the training time. 
In order to prevent tail classifiers from being over-suppressed, Equalization Loss proposes to ignore all negative gradients from head classes. Balanced Group Softmax (BAGS)~\cite{bags} puts categories with similar numbers of training instances into the same group and computes group-wise softmax cross-entropy loss respectively. BAGS achieves relative balance within each group, thus can effectively ameliorate the domination of the head classes over tail classes. 

The above methods can efficaciously reduce the suppression on tail classifiers. However, they need to partition the categories into several groups based on their category frequency prior. Such hard division between head and tail classes brings two problems, namely training inconsistency between adjacent categories and lack of discriminative power for rare categories. As shown in Figure~\ref{fig:distribution}, when two categories with similar instance statistics are divided into two different groups, there exists a huge gap between their training strategies. Such training inconsistency may deteriorate the network's performance. In general, it is suboptimal to distinguish the head and tail classes by the absolute number of samples. In addition, it frequently happens that two categories with high appearance similarity hold totally different sample frequency for datasets with large category vocabulary. For instance, category ``sunglasses" and ``eye mask" belong to the head and tail classes, respectively. To prevent the tail classifier ``eye mask" from being over-suppressed, the negative gradients from category ``sunglasses" is ignored by the classifier of ``eye mask". Nevertheless, this can also reduce the differentiability between these semantically similar cases in feature space, the classifier of ``eye mask" has a high chance to misclassify sample ``sunglasses" as ``eye mask". For those categories which belong to different groups but have high appearance similarity, it is hard for the network to learn a discriminative feature representation. 

Therefore, in this paper, we present a novel Adaptive Class Suppression loss (ACSL) to address the above two problems. The design philosophy is simple and straightforward: \emph{We assume all categories are from ``tail" group, and adaptively generate suppression gradients for each category according to their current learning status.} To be specific, we propose to treat all object categories as scarce categories, regardless of the statistics of per-category instances, thus eliminating the dilemma of manually defining the head and tail. Furthermore, in order to alleviate the insufficient learning and representation, we introduce an Adaptive Class Suppression Loss to adaptively balance the negative gradients between different categories, which effectively boosts the discriminative ability for tail classifiers. On the one hand, the proposed method can exterminate several heuristics and hyper-parameters of data distribution. On the other hand,  it can also avoid the problem caused by over-sampling and under-sampling, and ensure the training consistency for all classes and sufficient learning of rare or similar categories. Finally, it yields reliable and significant improvements in detection performance on large-scale benchmarks like LVIS and Open Images.

To sum up, this work makes the following three contributions:

\begin{enumerate}
	\item We propose a new statistic-free perspective to understand the long-tail distribution, thus significantly avoiding the dilemma of manual hard division. 
	\item We present a novel adaptive class suppression loss (ACSL) that can effectively prevent the training inconsistency of adjacent categories and improve the discriminative power of rare categories.
	\item We conduct comprehensive experiments on long-tail object detection datasets LVIS and Open Images. ACSL achieves 5.18\% and 5.2\% improvements with ResNet50-FPN on LVIS and OpenImages respectively, which validates its effectiveness.
\end{enumerate}

\section{Related Works}
\paragraph{General Object Detection} Current CNN-based object detectors can be divided into anchor-based and anchor-free detectors based on whether they depend on the anchor heuristic. 
Classic anchor-based detectors consist of one-stage and two-stage approaches. Two-stage detectors~\cite{ren2017faster,fast-r-cnn,dai2016rfcn,cai2018cascade,qin2019thundernet,He_2019_Bounding,Lin_2017_FPN,Pang_2019_libra} first generate coarse object candidates by a Region Proposal Network (RPN). And then the region features of these proposals will be extracted for accurate classification and bounding box regression. One-stage detectors~\cite{ssd,zhou2018scale,lin2020focal,chen2019towards,zhang2019freeanchor,zhang2018single} have a more concise structure. They make predictions on multiple feature maps directly without proposal generation process, thus enjoying higher computational efficiency.

Anchor-free pipelines abandon the anchor mechanism. They first locate several predefined or self-learned keypoints and then group them to final detections. CornerNet~\cite{cornernet,cornernet_light}, ExtremeNet~\cite{extremenet} and CenterNet~\cite{duan2019centernet} \etc represent one object as several predefined keypoints and detect by predicting these keypoints. RepPoints~\cite{yang2019reppoints} proposes a finer representation of objects as a set of sample points and learns to arrange them in a manner that bounds the spatial extent of an object. FCOS~\cite{tian2019fcos} solves object detection in a per-pixel prediction fashion. It predicts four distances for each location and filters those unreliable detections with novel centerness scores. They all perform well on balanced datasets. However, directly applying them to long-tail datasets achieves inferior performance due to the imbalanced data distribution. Thus, our intention is to improve the detectors' performance on long-tail datasets. 

\paragraph{Long-tail Recognition}
Re-sampling strategy is a typical technique for imbalanced datasets. Class-aware sampling~\cite{relay} and Repeat factor sampling~\cite{lvis} share the same design philosophy that they aim to balance the data distribution by adopting different sampling frequencies for different categories. Re-weighting~\cite{wang2017learning,LearningDeepRepresentation} is another widely used method which works by assigning weights for different training samples to emphasize the training of tail samples. Cui \etal~\cite{cui2019class} further proposes Class balanced loss which first calculates the effective number of samples for each category and uses the effective numbers to re-balance the loss. 

In addition, some works design specialized loss functions or training strategies to tackle the imbalanced issue. Equalization Loss~\cite{eql} ignores the negative gradients from head samples to prevent the tail classifiers from being over-suppressed. Cao \etal~\cite{ldam} proposes label-distribution-aware margin loss motivated by minimizing a margin-based generalization bound. Balanced Group Softmax~\cite{bags} puts categories with similar sample numbers to one group and applies softmax within each group. Balanced Meta-Softmax~\cite{ren2020balanced} devises an unbiased extension of softmax to accommodate the label distribution shift between training and testing. SimCal~\cite{simcal} proposes a simple calibration framework to more effectively alleviate classification head bias with a bi-level class balanced sampling approach. Kang \etal~\cite{kang2020decoupling} design a decoupling training schema, which first learns the representations and classifier jointly, then obtains a balanced classifier by re-training the classifier with class-balanced sampling.

The above methods can relieve the imbalanced training to a great extent. However, they all depend on the class distribution which brings inconvenience when we apply them to new long-tail datasets. In this paper, we devise a more general loss function which does not rely on the category frequency prior and can also handle the long-tail datasets well.

\section{Adaptive Class Suppression Loss}
In this section, we start by analyzing the limitations of group-based methods. Then, we introduce our proposed Adaptive Class Suppression Loss in detail and summary its advantages compared with previous methods.

\subsection{Limitations of Group-based Methods}
To pursue a balanced training for tail classifiers, several works propose to divide the whole categories into several groups based on the category frequency prior and adopt different training strategies for different groups. For instance, Equalization Loss~\cite{eql} ignores the negative gradients from samples of head categories to prevent the tail classifiers from being dominated by negative signals. Balanced Group Softmax~\cite{bags} first puts categories with similar sample numbers into the same group and applies softmax cross-entropy loss within each group. 

\begin{table}
	\caption{Experiments on LVIS with different groups. }
	\vspace{-0.5cm}
	\begin{center}
		\small
		\begin{tabular}{c|c|ccc}
			\specialrule{0.1em}{3pt}{3pt}
			Groups & $mAP$ &  $AP_r$  &  $AP_c$ &  $AP_f$ \\
			\specialrule{0.07em}{2pt}{2pt}
			(0,5)[5,$\infty$)      & 22.74 & 6.83  & 22.14 & 29.83 \\
			(0,50)[50,$\infty$)    & 25.30 & 15.11 & 24.99 & 29.77 \\
			(0,500)[500,$\infty$)  & 25.66 & 13.19 & 25.98 & 30.25 \\
			(0,5000)[5000,$\infty$)& 23.89 & 8.27  & 23.87 & 30.16 \\
			\specialrule{0.1em}{3pt}{3pt}
		\end{tabular}
	\end{center}
	\label{table:groups}
	\vspace{-0.7cm}
\end{table}

The group-based methods show their ability to improve the performance of tail classifiers. For these methods, how to properly divide all categories into different groups is of vital importance. We empirically find that the quality of grouping directly impacts the performance. We choose BAGS (one of the state-of-the-art group-based methods for long-tail object detection) with ResNet50-FPN as an example to illustrate the importance of proper grouping in LVIS dataset. For clarity, we divide all categories into 2 groups rather than the default 4 groups. We utilize different group partition strategies to train the detectors and report the $mAP$, $AP_r$ ($r$ for rare categories), $AP_c$ ($c$ for common) and $AP_f$ ($f$ for frequent) in Table~\ref{table:groups}. As we can see from this table, the dividing line need to be settled in a proper range (50\textasciitilde500) to achieve satisfactory performance. Setting it to a too large or small number all deteriorate the performance. 

\begin{figure}
	\centering
	\includegraphics[width=7.5cm]{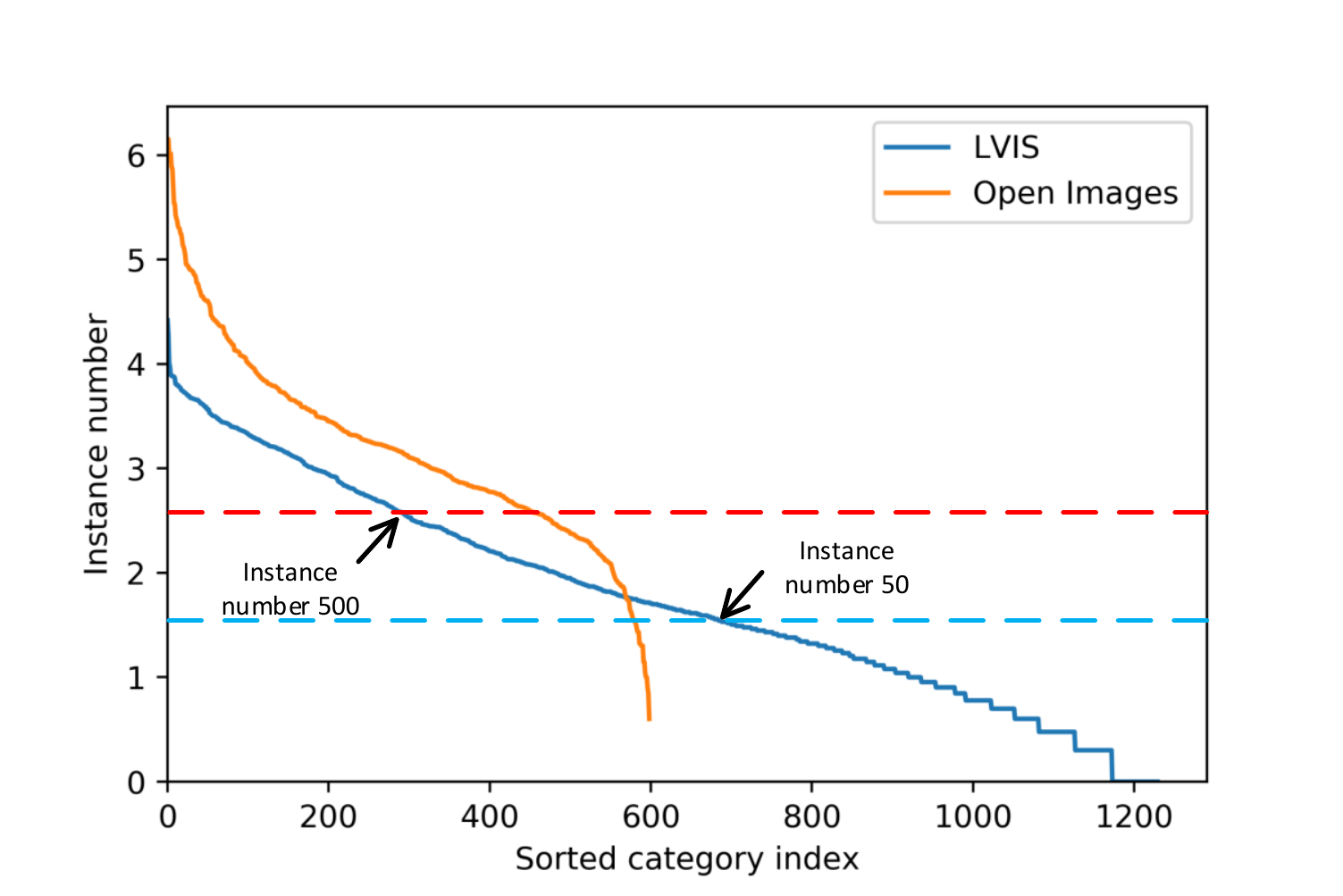}\\
	\caption{The data distribution of LVIS and Open Images dataset. The x-axis represents the sorted category index. Y-axis is the base-10 logarithm of the instance number.}
	\label{fig:datasets}
\end{figure}

Since the proper grouping is the precondition of good performance, it is necessary to divide groups based on the data distribution when applying group-based methods to other long-tail datasets. Nevertheless, different datasets usually have various data distributions. A good group partition for one dataset may be suboptimal for other datasets. As shown in Figure~\ref{fig:datasets}, models can obtain the best performance when the dividing line is set to 500 in LVIS. However, it is not a good partition for Open Images dataset since it has fewer categories and each category has more samples. Finding a proper group partition strategy for a new dataset could be laborious and time-consuming, which will limit the application of group-based methods in real-world scenarios. We naturally ask a question: can we devise a more general method that can be directly applied to other long-tail datasets without extra heuristics? To accomplish this goal, we propose a novel Adaptive Class Suppression Loss which does not need to perform group partition nor depend on the category frequency prior. It can be easily combined with different datasets seamlessly.

\begin{figure}
	\centering
	\includegraphics[width=8cm]{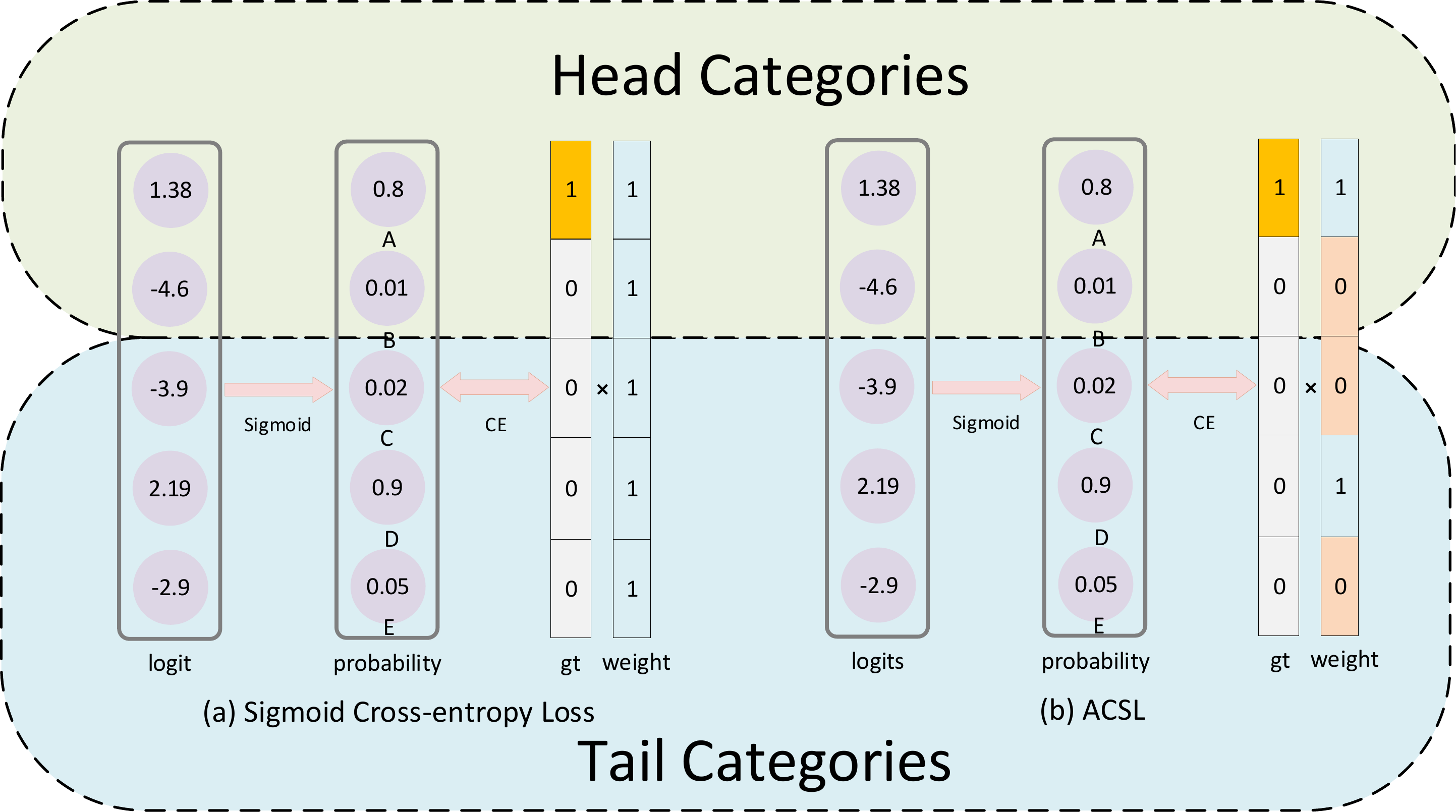}\\
	\caption{An illustration of Sigmoid Cross-entropy Loss and our proposed ACSL. The top two classes belong to head categories and the bottom three classes belong to tail categories. For ACSL, the hyper-parameter $\xi$ is 0.7. }
	\label{fig:acsl}
	\vspace{-0.5cm}
\end{figure}

\subsection{Formulation of ACSL}\label{sec:acsl}

Based on the above analysis, we believe a satisfactory loss function for long-tail datasets should have the following two properties: 

\begin{enumerate}[(i)]
\item {\emph{The tail classifiers should be protected from being over-suppressed by the overwhelming head samples. In the meantime, negative gradients for easily confused categories should be preserved for discriminative learning.}}

\item {\emph{The loss function should better not depend on the label distribution prior so that it can be seamlessly applied to diverse datasets without re-calculating the new category distribution statistics. }}
\end{enumerate}

Giving a sample $x_s$ belonging to the head category $k$, its label is a one-hot vector $Y$, in which $y_k$ equals to 1 and $y_i (i \neq k)$ is set to 0. $z_i$ represents the network output logit of category $i$. By applying the sigmoid function to the logit $z_i$ (Equation~(\ref{eq:sigmoid})), we can obtain $p_i$, the probability of current sample belonging to category $i$. The binary cross-entropy loss is formulated as Equation~(\ref{eq:bce}). Then we can derive the gradient of the loss function with respect to logit $z_i$ as in Equation~(\ref{eq:bce-grad}). For tail category $i \neq k$, BCE generates negative suppression gradients to force the classifier $i$ to output low confidence. Such suppression gradients is beneficial to some extent. However, the excessive suppression gradients derived from the head categories will seriously hinder the positive activation of tail categories.

\begin{equation}
\label{eq:sigmoid}
p_i = \frac{1}{1+e^{-z_i}}
\end{equation}

\begin{equation}
\label{eq:bce}
L_{BCE}(x_s)=-\sum_{i=1}^{C}{log(\hat{p_i})}
\end{equation}

where, 
\begin{equation}
\label{eq:prob}
\resizebox{0.45\linewidth}{!}{$
	\displaystyle
	\hat{p_i}=\left\{
	\begin{aligned}
	& \quad p_i, \quad\quad\text{if $i = k$} \\
	&1-p_i, \quad\,\,\text{if $i \neq k$}\\
	\end{aligned}
	\right.
	$}
\end{equation}

\begin{equation}
\label{eq:bce-grad}
\resizebox{0.57\linewidth}{!}{$
	\displaystyle
	\frac{\partial L_{BCE}}{\partial z_i}=\left\{
	\begin{aligned}
	&  p_i-1, \quad\,\,\text{if $i = k$} \\
	& p_i, \quad\quad\quad\,\text{if $i \neq k$}\\
	\end{aligned}
	\right.
	$}
\end{equation}

Therefore, we devise a novel Adaptive Class Suppression Loss to protect the training of tail classifiers and adaptively choose which categories should be suppressed based on the learning status. As shown in Equation~(\ref{eq:acsl}), we multiply a binary weight term $w_i$ to the loss term $-log(\hat{p_i})$ for category $i$. For category $k$, $w_i$ is set to 1 since the current sample belongs to category $k$. For other categories $i(i \neq k)$, $w_i$ controls whether the network applies suppression on category $i$. Here, we utilize the output confidence $p_i$ as a signal to determine whether to suppress category $i$. If $p_i$ is larger than the predefined threshold $\xi$, that means the network is confused between category $i$ and $k$. Hence, we set $w_i$ to 1 to perform discriminative learning. Otherwise, $w_i$ will be set to 0 to avoid numerous unnecessary negative suppression. Rather than depending on the category distribution statistics, our proposed ACSL only relies on the network output confidences which saves the efforts of finding optimal category statistics related hyper-parameters when switching to a new dataset. The formulation is defined as follows:

\begin{equation}
\label{eq:acsl}
L_{ACSL}(x_s)=-\sum_{i=1}^{C}{w_i log(\hat{p_i})}
\end{equation}

where,
\begin{equation}
\label{eq:wi}
\resizebox{0.55\linewidth}{!}{$
	\displaystyle
	w_i=\left\{
	\begin{aligned}
	& \quad 1, \quad\quad\quad\text{if $i = k$} \\
 	& \quad 1, \quad\text{if $i \neq k$ and $p_i \geq \xi$}\\
	& \quad 0, \quad\text{if $i \neq k$ and $p_i < \xi$}
	\end{aligned}
	\right.
	$}
\end{equation}

The gradient of the loss function with respect to $z_i$ can be derived as Equation~(\ref{eq:sigmoid-grad}). 

\begin{equation}
\label{eq:sigmoid-grad}
\resizebox{0.57\linewidth}{!}{$
	\displaystyle
	\frac{\partial L_{ACSL}}{\partial z_i}=\left\{
	\begin{aligned}
	&  p_i-1, \quad\quad\,\,\text{if $i = k$} \\
	& w_ip_i, \quad\quad\quad\,\text{if $i \neq k$}\\
	\end{aligned}
	\right.
	$}
\end{equation}

As shown in Figure~\ref{fig:acsl}, we give a simple illustration to better understand how ACSL works. For sigmoid cross-entropy loss (Figure~\ref{fig:acsl}(a)), it does not consider the imbalanced class distribution, and directly generate negative gradients for all other categories (the weight vector is filled with 1), thus leading to the tail categories receiving a large number of suppression information continuously, which severely reduces the discrimination ability of tail classifiers. For the proposed ACSL, the network can adaptively generate suppression gradients for tail categories. As show in Figure~\ref{fig:acsl}(b), the network simultaneously yields high confidence 0.9 for category ``D", which means the category ``A" is semantically similar to the category ``D". Therefore, it is necessary to generate negative suppression for category ``D", but not for the remaining tail categories with low confidences.


\subsection{Advantages over Previous Methods}
Compared with previous methods, the proposed ACSL has the following three characters that makes it more suitable for the long-tail datasets: 

\emph{ACSL takes the network learning status into consideration.} Previous methods~\cite{class-imbalance,relay,eql,bags} only consider the label distribution, while ignore the network learning status. For instance, state-of-the-art Equalization Loss~\cite{eql} calculates a binary weight for each category based on the sample numbers. Once the dataset is given, the weight will be determined and will not be changed during the whole training process. ACSL adaptively choose which categories to suppress based on the output confidences, which takes the network learning status into consideration, resulting in a more efficient learning process. 

\emph{ACSL works in a more fine-grained sample level.} Previous methods perform an identical operation on samples from the same category. For example, Class balanced loss~\cite{class-imbalance} assigns the same weights for samples from one category. Equalization loss~\cite{eql} generates the same weight masks for two samples if they all belong to the same category. These methods all ignore the diversity of variant samples. In contrast, ACSL calculates the category weights for a sample based on its output confidences. The generated category weights for samples from the same category might be different. Thus, ACSL can control the training of each classifier in a more accurate way.  

\emph{ACSL does not depend on the class distribution.} The label distribution is indispensable for previous methods. They need to know the class distribution in advance to design sampling strategy~\cite{relay,lvis}, determine the weights for samples~\cite{class-imbalance} or divide categories into groups~\cite{bags,eql}. This is inefficient since the strategies must be re-designed based on the new data distribution when we want to apply these methods to new long-tail datasets. In comparison, ACSL does not require the category frequency prior, which means it can be applied to new long-tail datasets seamlessly. We empirically find that ACSL works well on LVIS and Open Images under the same hyper-parameter setting. 
 
\begin{table}
	\caption{Experimental results of ACSL with different $\xi$. }
	\vspace{-0.5cm}
	\begin{center}
		\small
		\begin{tabular}{c|c|c|ccc}
			\specialrule{0.1em}{3pt}{3pt}
			& $\xi$ &  $mAP$  &  $AP_r$  &  $AP_c$  &  $AP_f$   \\
			\specialrule{0.1em}{3pt}{3pt}
			baseline (1x)        &  $-$  &  21.18  &  4.30  &  20.09  &  29.28  \\
			baseline (2x)        &  $-$  &  22.28  &  7.38  &  22.34  &  28.17  \\
			\specialrule{0.07em}{2pt}{2pt}
			\multirow{6}*{ACSL}  & 0.01  &  23.53  &  11.48 &  22.73  &  29.35  \\
			&  0.1  &  25.11  &  16.04 &  24.72  &  29.22  \\
			&  0.3  &  25.72  &  17.65 &  25.45  &  29.27  \\
			&  0.5  &  26.08  &  18.61 &  25.85  &  29.36  \\
			&  0.7  &  \textbf{26.36}  &  \textbf{18.64} &  \textbf{26.41}  &  29.37  \\
			&  0.9  &  25.99  &  17.25 &  26.0   &  \textbf{29.46}  \\
			\specialrule{0.1em}{3pt}{3pt}
		\end{tabular}
	\end{center}
	\label{table:ablation}
	\vspace{-0.5cm}
\end{table}

\begin{table}
	\caption{Results with larger backbones ResNet101, ResNeXt-101-64x4d and stronger detector Cascade R-CNN.}
	\vspace{-0.5cm}
	\begin{center}
		\small
		\begin{tabular}{c|c|c|ccc}
			\specialrule{0.1em}{3pt}{3pt}
			Models                      &     Method    &  $mAP$ & $AP_r$ & $AP_c$ & $AP_f$ \\
			\specialrule{0.07em}{2pt}{2pt}
			\multirow{2}*{Faster R101}  &   baseline    & 22.36  & 3.14 & 21.82 & \textbf{30.72} \\
			&     Ours      &  \textbf{27.49}  &  \textbf{19.25}  &  \textbf{27.60}  &  30.65  \\
			\specialrule{0.07em}{2pt}{2pt}
			\multirow{2}*{Faster X101}  &    baseline   & 24.70  &  5.97  & 24.64  & \textbf{32.26}  \\
			&     Ours      & \textbf{28.93}  &  \textbf{21.78} & \textbf{28.98}  & 31.72  \\
			\specialrule{0.07em}{2pt}{2pt}
			\multirow{2}*{Cascade R101} &    baseline   &  25.14  &  3.96  &  24.55  &  \textbf{34.35}  \\
			&      Ours     &  \textbf{29.71}  &  \textbf{21.72}  &  \textbf{29.43}  &  33.26  \\
			\specialrule{0.07em}{2pt}{2pt}
			\multirow{2}*{Cascade X101} &    baseline   & 27.14  &  4.36  & 27.32  & \textbf{36.03}  \\
			&     Ours      & \textbf{31.47}  &  \textbf{23.39} & \textbf{31.50}  & 34.66  \\
			\specialrule{0.1em}{3pt}{3pt}
		\end{tabular}
	\end{center}
	\label{table:large-model}
	\vspace{-0.5cm}
\end{table}

\begin{figure*}[htbp]
	\centering
	\vspace{-0.15cm}
	\subfigure[The $AP$ on frequent categories]
	{
		\begin{minipage}{5.0cm}
			\centering
			\includegraphics[width=5.0cm]{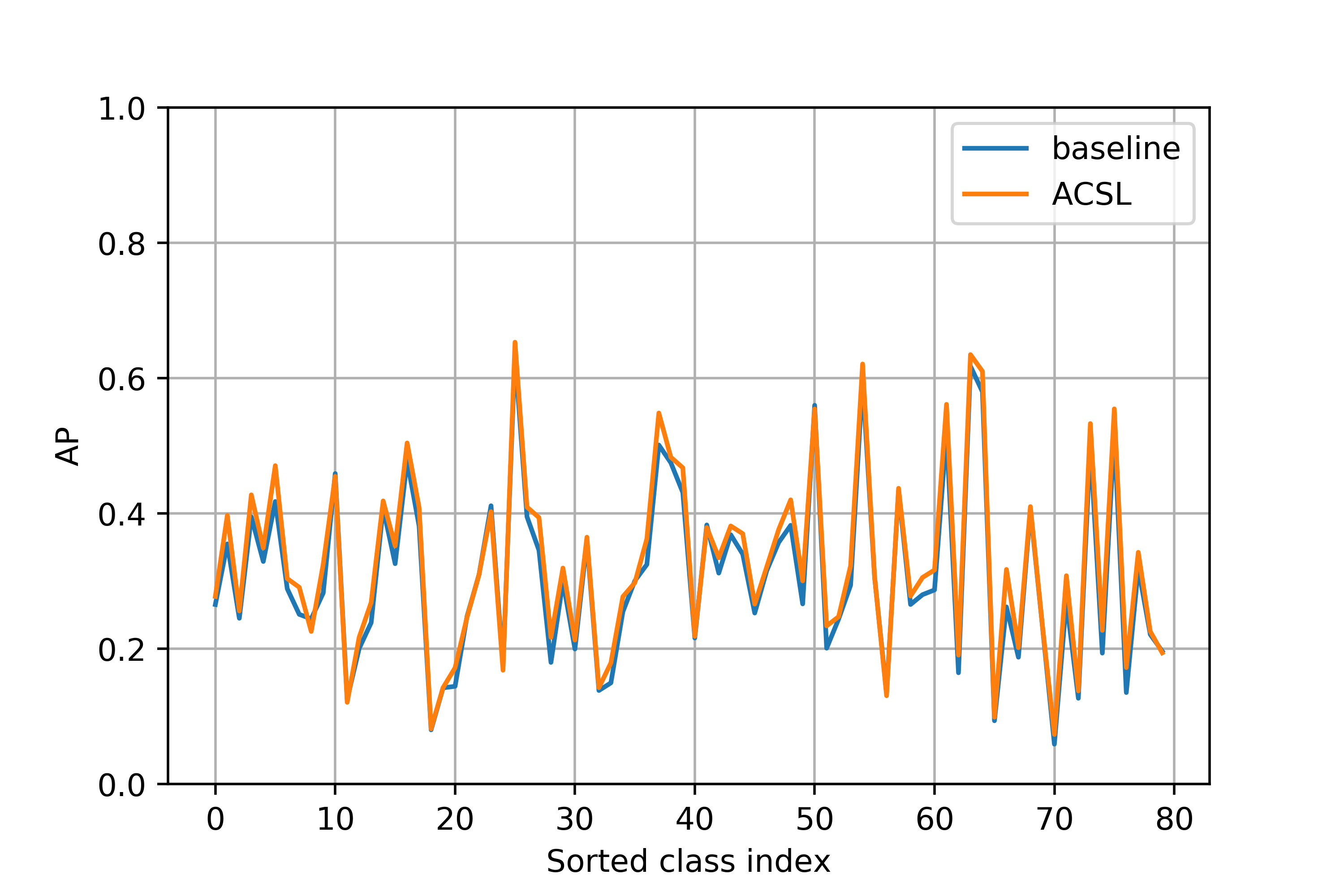}
			\label{fig:frequent_ap}
		\end{minipage}
	}
    \vspace{-0.15cm}
	\subfigure[The $AP$ on common categories]
	{
		\begin{minipage}{5.0cm}
			\centering
			\includegraphics[width=5.0cm]{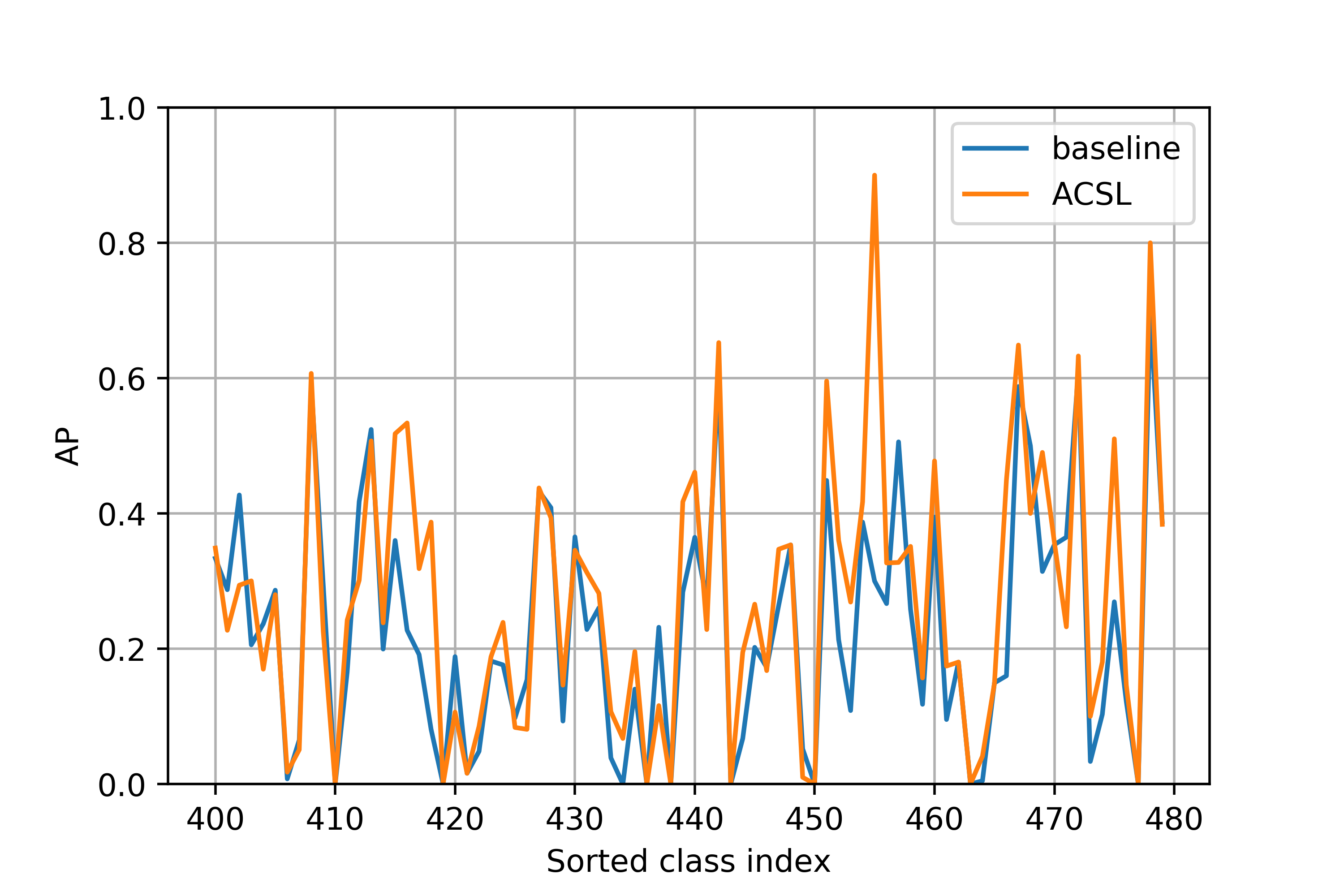}
			\label{fig:common_ap}
		\end{minipage}
	}
    \vspace{-0.15cm}
	\subfigure[The $AP$ on rare categories]
	{
		\begin{minipage}{5.0cm}
			\centering
			\includegraphics[width=5.0cm]{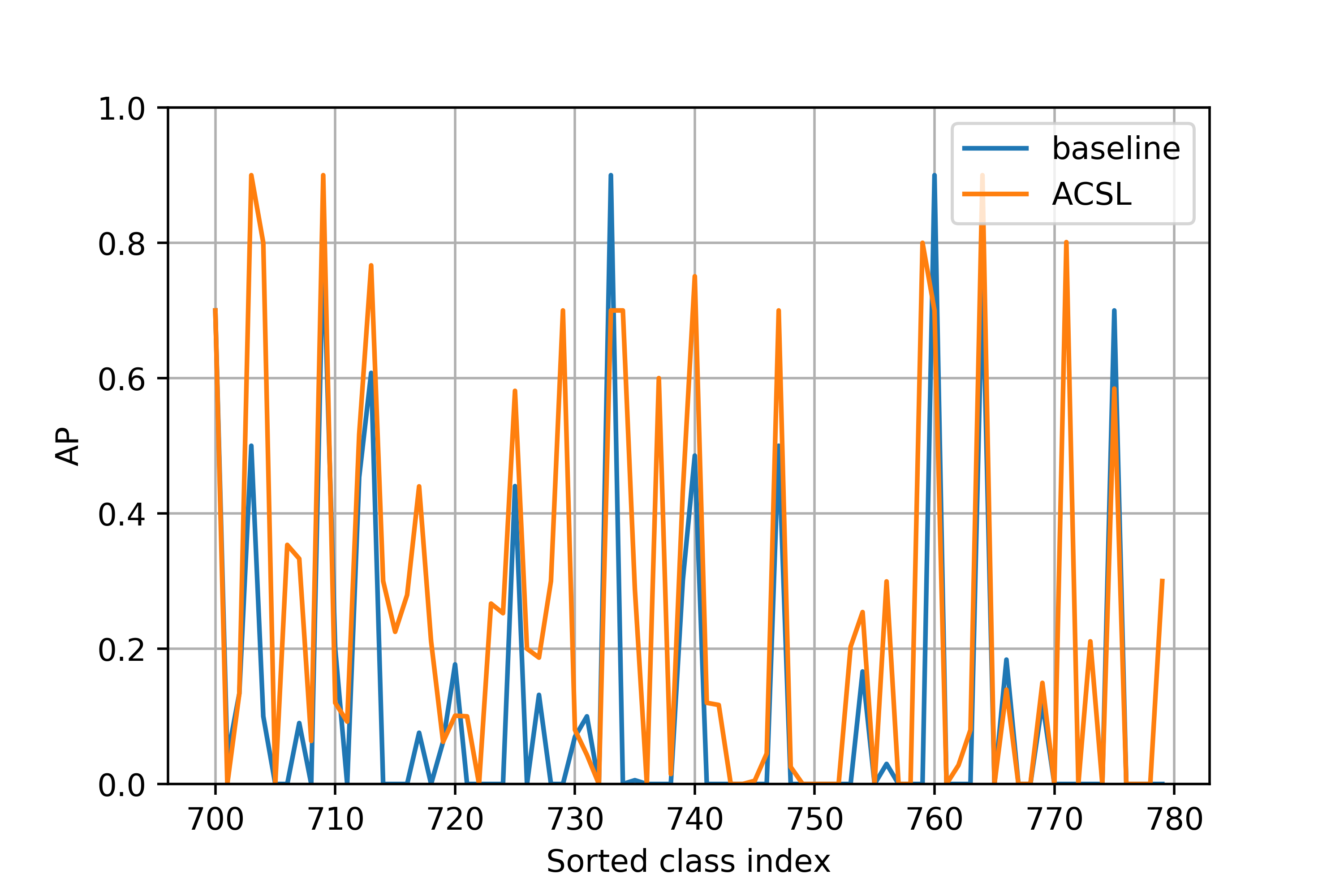}
			\label{fig:rare_ap}
		\end{minipage}
	}
	\caption{The $AP$ of baseline and ACSL on frequent, common and rare categories, respectively. Both models are trained with ResNet50-FPN backbone. The x-axis is the sorted class index. The y-axis means the precision. }
	\label{fig:cls_ap}
	\vspace{-0.3cm}
\end{figure*}

\section{Experiments on LVIS}

\subsection{Dataset and Setup}
To validate the effectiveness of the proposed ACSL, we conduct comprehensive experiments on the long-tail Large Vocabulary Instance Segmentation (LVIS) dataset~\cite{lvis}. In this work, we use the LVIS-v0.5, which contains 1230 categories with both bounding box and instance mask annotations. We train the models with 57k \emph{train} images and report the accuracy on 5k \emph{val} images. LVIS divides all categories into 3 groups based on the number of images that contain those categories: frequent ($>$100 images), common (11-100 images) and rare ($<$10 images). We use the official evaluation API\footnote{ https://github.com/lvis-dataset/lvis-api} to evaluate all the models. Besides the official metrics $mAP$, the $AP_r$ (AP for rare classes), $AP_c$ (AP for common classes) and $AP_f$ (AP for frequent classes) are also reported. 

\subsection{Implementation Details}
We choose the classic two-stage detector Faster R-CNN with FPN structure as the baseline. The training image is resized such that the shorter edge is 800 pixels and the longer edge is no more than 1333. When training, we use 8 GPUs with a total batch size 16. The optimizer is stochastic gradient descent (SGD) with momentum 0.9 and weight decay 0.0001. The initial learning rate is 0.02 with 500 iterations' warm up. For 1x training schedule, learning rate decays to 0.002 and 0.0002 at epoch 8 and 11, respectively. The training stops at the end of epoch 12. When testing, we first apply Non-Maximum Suppression with IoU threshold 0.5 to remove duplicates. Then, the top 300 detections will be selected as the final results. Other hyper-parameter settings like anchor scales and anchor ratios follow the default settings in MMDetection~\cite{mmdetection}. 

When we conduct experiments with ACSL or other specialized loss functions designed for long-tail datasets, we focus on the classification subnet of Faster R-CNN and replace the softmax cross-entropy loss with these specialized loss functions. Inspired by~\cite{kang2020decoupling}, we decouple the training of feature representation and the classifier. We train a naive Faster R-CNN detector in the first stage and fine-tune the classifier in the second stage. In object detection, the background samples are negative samples for all categories. We adopt a simple strategy to chase a relative balance for each classifier. Since the ratio between the image numbers of rare, common and frequent categories is approximately 1:10:100, we randomly choose 1$\%$, 10$\%$ background samples for rare and common categories, respectively.
 
\begin{table*}
	\caption{Comparison with state-of-the-art methods on LVIS-v0.5 \emph{val} dataset. \textbf{Bold} numbers denote the best results among all models. ``ms" means multi-scale testing. }
	\vspace{-0.6cm}
	\begin{center}
		\small
		\begin{tabular}{c|c|c|ccc|cc|ccc}
			\specialrule{0.1em}{3pt}{3pt}
			Methods      &    backbone    & $mAP$ & $AP_r$ & $AP_c$ & $AP_f$ & $AP@0.5$ & $AP@0.75$ & $AP_s$ & $AP_m$ & $AP_l$ \\
			\specialrule{0.07em}{2pt}{2pt}
			Focal Loss~\cite{lin2020focal}  &  \multirow{6}*{ResNet-50}  & 21.95 &  10.49 & 22.42  & 25.93  &   35.15   &   23.91   & 18.66  & 28.59  & 31.46  \\ 
			CBL~\cite{cui2019class} &        & 23.9  &  11.4  & 23.8   & 27.3   &   $-$     &   $-$     &  $-$   &  $-$   &  $-$  \\ 
			LDAM~\cite{ldam}        &                & 24.5  &  14.6  & 25.3   & 26.3   &   $-$     &   $-$     &  $-$   &  $-$   &  $-$  \\ 
			RFS~\cite{lvis}         &                & 24.9  &  14.4  & 24.5   & 29.5   &   41.6    &   25.8    &  19.8  & 30.6   & 37.2   \\
			LWS~\cite{kang2020decoupling}         &                & 24.1  &  14.4  & 24.4   & 26.8   &   $-$     &   $-$     &  $-$   &  $-$   &  $-$  \\ 
			SimCal~\cite{simcal}       &                & 23.4  &  16.4  & 22.5   & 27.2   &   $-$     &   $-$     &  $-$   &  $-$   &  $-$  \\ 
			\specialrule{0.07em}{2pt}{2pt}
			\multirow{3}*{EQL~\cite{eql}}  &       ResNet-50         & 25.06 &  11.92 & 25.98  & 29.14  &   40.14   &   27.30   & 20.08  & 31.50  & 38.67  \\
			& ResNet-101 &  26.05  &  11.45  &  27.14  &  30.51  &  41.30  &  27.83  &  20.35  &  33.73  &  40.75 \\
			& ResNeXt-101-64x4d &  28.04  &  15.03  &  29.14  &  31.87  &  44.06  &  30.07  &  22.19  &  34.52  &   42.97  \\
			\specialrule{0.07em}{2pt}{2pt}
			\multirow{3}*{BAGS~\cite{bags}}        &      ResNet-50          & 25.96 &  17.65 & 25.75  & 29.54  &   43.58   &   27.15   & 20.26  & 32.81  & 40.10  \\
			&   ResNet-101 &  26.39  &  16.80  &  25.82  &  30.93  &  43.44  &  27.63  &  20.29  &  34.39  &  41.07 \\
			&  ResNeXt-101-64x4d &  27.83  &  18.78  &  27.32  &  32.07  &  45.83  &  28.99  &  21.92  &  35.65  &   43.11  \\
			\specialrule{0.07em}{2pt}{2pt}
			\multirow{6}*{ACSL (Ours)} &    ResNet-50    & 26.36 &  18.64 & 26.41 & 29.37  &    42.38   &   28.63   & 20.43  & 33.11  & 40.21  \\
			&   ResNet-101 &  27.49  &  19.25  &  27.60  &  30.65  &  43.45  &  29.69  &  21.11  &  34.96  &  42.00\\
			&  ResNeXt-101-64x4d &  28.93  &  \textbf{21.78}  &  28.98  &  31.72  &  45.54  &  31.19  &  22.16  &  35.81  &  43.43   \\
			&    ResNet-50 (ms)    & 27.24 & 17.86 & 27.42 & 30.76 & 44.46 & 28.54 & 20.96 & 34.40 & 41.68 \\
			
			&   ResNet-101 (ms) & 28.23 & 17.42 & 28.40 & 32.32 & 44.73 & 30.13 & 21.86 & 35.43 & 44.06 \\
			
			&  ResNeXt-101-64x4d (ms) & \textbf{29.47} & 20.30 & \textbf{29.45} & \textbf{33.15} & \textbf{46.82} & \textbf{31.55} & \textbf{22.52} & \textbf{37.32} & \textbf{45.51} \\
			\specialrule{0.1em}{3pt}{3pt}
		\end{tabular}
	\end{center}
	\label{table:sota}
	\vspace{-0.5cm}
\end{table*}

\subsection{Ablation Study}
We take the Faster R-CNN with ResNet50-FPN backbone as the baseline model. As we first train a naive Faster R-CNN detector for 12 epochs in the first stage and then fine-tune the classifier with ACSL for another 12 epochs, the total training epoch is 24. We also report the results of the baseline model trained with 24 epochs for a fair comparison. As the results in Table~\ref{table:ablation}, the baseline (1x) model has a relatively high precision on frequent categories but an extremely low accuracy on rare categories, merely 4.3\%. More training iterations bring benefits to rare and common categories and lift the overall $mAP$ by 1.1\%. Even so, the performance of rare categories is still unsatisfactory, which shows the bottleneck of the baseline model lies in the tail classes. 

ACSL introduces a hyper-parameter $\xi$ to define the easily confused categories. It is a trade-off between reliving over-suppression on tail classes and chasing discriminative learning. A small $\xi$ means that most of the categories will be suppressed, which will suppress too much on tail categories. However, for an extremely large $\xi$, the network will only suppress categories with extremely high confidences while ignore most of the other categories, thus will weaken the classifier's discriminative power. To explore how $\xi$ influences the performance, we conduct experiments with several different values and report the results in Table~\ref{table:ablation}. As shown in this table, setting $\xi$ to 0.01 improves the precision but the improvements are limited. As $\xi$ grows larger, the $mAP$ and $AP_r$ also increase which proves that properly increasing $\xi$ relieves the suppression on tail classifiers. When $\xi$ increases to 0.7, the model achieves the best performance. The overall $mAP$ is 26.36\%, surpassing the baseline model by considerable 5.18\%. Nevertheless, continuing to increase its value deteriorates the performance since it will weaken the discriminative power of the classifier. We empirically find $\xi=0.7$ works best under current setting. 

\subsection{Generalization on Stronger Models}
To verify the generalization of our method, we further conduct experiments on stronger backbones and detectors. We replace the ResNet50 backbone with larger ResNet101 and ResNeXt-101-64x4d. The results are summarized in Table~\ref{table:large-model}. Experimental results reveal that ACSL can still achieve competitive results on larger backbones. With ResNet101 backbone, ACSL outperforms the baseline model by 5.13\% $mAP$. And we observe that ACSL brings significant performance gains for rare categories with various backbones (16.11\% $AP_r$ improvements for ResNet101, for instance), which demonstrates the advantage of ACSL when tackling with long-tail datasets. The advantage of ACSL still exists when we use a larger backbone ResNeXt-101. The $mAP$ of ACSL is 4.23\% higher than the baseline model for ResNeXt-101. Moreover, the utilization of ACSL is not limited to a certain type of detector. It can be easily combined with other detectors such as Cascade R-CNN. For Cascade R-CNN, we replace the softmax cross-entropy loss with ACSL on all 3 heads and fine-tune the classifiers' weights in the second training stage. With ResNet-101 Cascade R-CNN detector, ACSL achieves 29.71\% $mAP$, surpassing baseline model by 4.57\% $mAP$. When using larger backbone ResNeXt-101, the overall $mAP$ can be further pushed to 31.47\%, outperforming the baseline by a significant 4.33\%. It is worth noticing that ACSL obtains the best performance on the overall $mAP$, $AP_r$ and $AP_c$, which further proves ACSL's ability to tackle long-tail datasets. 

\subsection{Performance Analysis}
In order to have a more intuitive sense of how ACSL influences the network's performance, we visualize the performance of baseline and ACSL on different categories in Figure~\ref{fig:cls_ap}. Figure~\ref{fig:frequent_ap} shows the $AP$ on frequent categories. The two curves almost overlap with each other which demonstrates that ACSL does not harm the training of head classifiers. For common categories (Figure~\ref{fig:common_ap}), ACSL begins to show its advantages over baseline model. The precision curve of ACSL almost covers that of the baseline model, which shows that ACSL brings moderate improvements for common categories. As the decreasing of categories' sample numbers, the advantages of ACSL becomes more significant. As showed in Figure~\ref{fig:rare_ap}, ACSL outperforms baseline by a landslide. The integral area of the orange curve (ACSL) is much larger than that of the blue one (baseline). These three figures indicate that ACSL is able to improve the performance of tail categories without sacrificing the precision of head classes. 

\subsection{Comparison with State of the Arts}
In this section, we compare the performance of the proposed ACSL with other state-of-the-art methods and report the results in Table~\ref{table:sota}. All the models adopt Faster R-CNN with FPN structure. As shown in the table, without any bells and whistles, our single ResNet-50 model achieves 26.36\% $mAP$. It surpasses other competitive methods, including Equalization Loss (25.06\% $mAP$) and BAGS (25.96\% $mAP$). And we do observe that the performance gain comes from the tail classes (rare and common categories). Among all the single ResNet-50 models, ACSL obtains the best performance on rare (18.64\%) and common categories (26.41\%). In the meantime, it does not sacrifice the precision of head categories in exchange for the improvements on tail classes. The observations above exhibit the ability of ACSL to tackle the class imbalance problem in long-tail datasets. 

The advantages of ACSL becomes more significant when it is combined with larger backbones. With ResNet-101 backbone, ACSL achieves 27.49\% $mAP$, outperforming Equalization Loss and BAGS by 1.44\% and 1.1\% $mAP$, respectively. For stronger backbone ResNeXt-101-64x4d, ACSL obtains a 28.93\% $mAP$, surpassing the counterparts of Equalization Loss (28.04\% $mAP$) and BAGS (27.83\% $mAP$). When applying multi-scale testing strategy, the performance can be further improved. Finally, our best model achieves 29.47\% $mAP$.   

\begin{table}
	\caption{Experiments on Open Images with different backbones. }
	\vspace{-0.6cm}
	\begin{center}
		\small
		\begin{tabular}{ccc}
			\specialrule{0.1em}{3pt}{3pt}
			Backbone & Methods &  $AP$     \\
			\specialrule{0.07em}{2pt}{2pt}
			\multirow{2}*{ResNet50-FPN} & baseline & 55.1 \\
			& ours     & \textbf{60.3} \\
			\specialrule{0.07em}{2pt}{2pt}
			\multirow{2}*{ResNet101-FPN}& baseline & 56.3 \\
			& ours     & \textbf{61.6} \\
			\specialrule{0.07em}{2pt}{2pt}
			\multirow{2}*{ResNet152-FPN}& baseline & 57.4 \\
			& ours     & \textbf{62.8} \\ 
			\specialrule{0.1em}{3pt}{3pt}
		\end{tabular}
	\end{center}
	\label{table:openimage}
	\vspace{-0.6cm}
\end{table}

\section{Experiments on Open Images}
Open Images V5 is the largest existing object detection dataset which contains a total of 16M bounding boxes for 600 object classes on 1.9M images. For such a large scale dataset, it is hard to maintain a relatively balanced class distribution. In fact, it suffers from severe class imbalance. For instance, the most frequent category ``Man" contains 378077 images, which is 126 thousand times of the rarest category ``Paper cutter" (3 images). The whole dataset consists of three parts: \emph{train}, \emph{validation} and \emph{test}. All models are trained with 1.7M \emph{train} images and evaluated on \emph{validation} dataset. When performing evaluation, we use the official evaluation code for Open Images which evaluates only at IoU threshold 0.5. To reduce the training time on such a large dataset, we use the large batch training framework LargeDet~\cite{largebatch} to train our models. 

To testify the effectiveness of ACSL, we conduct experiments with various backbones including ResNet50, ResNet101 and ResNet152. The hyper-parameter $\xi$ of ACSL is set to 0.7, the same value as in LVIS. Since objects in Open Images have multiple labels, we train the models under multiple label setting. The experimental results are summarized in Table~\ref{table:openimage}. As shown in this table, the models trained with ACSL outperform the baseline models by a large margin. For ResNet50-FPN detector, ACSL achieves 60.3\% $AP$, 5.2\% higher than the baseline. For larger backbone ResNet152, our model obtains 62.8\% $AP$, surpassing the baseline by 5.4\% $AP$. To validate that the performance gains mainly come from the rare categories, we extract the precisions of some rare categories and list them in Table~\ref{table:oid_cls}. We can observe that ACSL achieves remarkable performance improvements for rare categories. For rare category ``Face powder" (``Fac" for short), the performance gain is 63.1\% $AP$. As shown in this table, ACSL brings considerable improvements to other rare categories, leading to higher overall precision.

\begin{table}
	\caption{The detailed precision on some of the tail categories of Open Images.}
	\vspace{-0.6cm}
	\begin{center}
		\scriptsize
		\setlength\tabcolsep{4.5pt}  
		\begin{tabular}{cccccc}
			\specialrule{0.1em}{3pt}{3pt}
			& Spa & Scr & Fac & Cas & Hor  \\
			\specialrule{0.07em}{2pt}{2pt} 
			img num & 38 &46 &49 &53 &54  \\
			\specialrule{0.07em}{2pt}{2pt}
			baseline & 35.0 &46.6 &17.8 &19.9 & 8.3   \\
			ACSL     & \textbf{41.6(+6.6)} & \textbf{55.6(+9.0)} & \textbf{80.9(+63.1)} & \textbf{47.5(+27.6)} & \textbf{16.6(+8.3)}  \\
			\specialrule{0.1em}{3pt}{3pt}
			& Slo & Obo & Squ & Bin & Ser \\
			\specialrule{0.07em}{2pt}{2pt} 
			img num &103 & 93 &97  &109  &106\\
			\specialrule{0.07em}{2pt}{2pt}
			baseline &25.0 & 22.2 &29.1  &42.7  &40.2\\
			ACSL     & \textbf{45.0(+20)} & \textbf{83.3(+61.1)} & \textbf{50.3(+21.2)}  & \textbf{61.5(+18.8)} & \textbf{73.2(+33)} \\
			\specialrule{0.1em}{3pt}{3pt}
		\end{tabular}
	\end{center}
	\label{table:oid_cls}
	\vspace{-0.6cm}
\end{table}

\begin{table}
	\caption{Comparison with other methods on Open Images. All models are trained with ResNet50-FPN backbone and evaluated on 500 categories. }
	\vspace{-0.6cm}
	\begin{center}
		\small
		\begin{tabular}{cc}
			\specialrule{0.1em}{3pt}{3pt}
			Method   &  AP    \\
			\specialrule{0.07em}{2pt}{2pt}
			Class Aware Sampling~\cite{relay}  & 56.50  \\
			Equalization Loss~\cite{eql}       & 57.83  \\
			Ours                    &   \textbf{61.70}  \\
			\specialrule{0.1em}{3pt}{3pt}
		\end{tabular}
	\end{center}
	\label{table:oid_others}
	\vspace{-0.9cm}
\end{table}

We also compare our method with other methods on Open Images dataset. The results are summarized in Table~\ref{table:oid_others}. From Table~\ref{table:oid_others}, we can observe the performance advantage of our method with the same backbone. And the effectiveness of our method can be adequately validated. 

\section{Conclusion}
In this work, we present a novel adaptive class suppression loss (ACSL) for long-tail datasets. ACSL is able to prevent the classifiers of tail categories from being over-suppressed by the samples from head categories. In the meantime, it preserves the discriminative power between easily confused categories. Equipped with ACSL, the models can achieve higher precision, especially on tail classes. Experimental results on long-tail datasets LVIS and Open Images validate its effectiveness. We will also apply the proposed ACSL to the long-tail image classification datasets in the future work.

\noindent \textbf{Acknowledgement}: This work was supported by the Research and Development Projects in the Key Areas of Guangdong Province (No.2019B010153001) and National Natural Science Foundation of China under Grants No.61772527, No.61976210, No.61876086, No.62002357, No.61806200, No.62002356, No.51975044 and No.61633002.

{\small
\bibliographystyle{ieee_fullname}
\bibliography{egbib}
}

\end{document}